\documentclass[runningheads]{llncs}

\usepackage[mobile]{eccv}

\usepackage{eccvabbrv}

\usepackage{graphicx}
\usepackage{booktabs}

\usepackage[accsupp]{axessibility}  

\usepackage[pagebackref,breaklinks,colorlinks,bookmarks=false,citecolor=eccvblue]{hyperref}

\usepackage{orcidlink}

\usepackage{subcaption}
\usepackage{amsmath}
\usepackage{amssymb}
\usepackage{dsfont}
\usepackage{bbding}    
\usepackage{xcolor}
\definecolor{mydarkgreen}{rgb}{0,.7,0}
\usepackage{url}
\usepackage{multirow}
\usepackage{listings}
\usepackage{tabularx} 

\usepackage{colortbl}  
\colorlet{mylightblue}{white!90!blue}
\colorlet{mylightgray}{white!90!gray}
\newcommand{\comment}[1]{}

\def\bfa#1{\textbf{\color{black}{#1}}}

\begin{document}

\title{OpenKD: Opening Prompt Diversity for Zero- and Few-shot Keypoint Detection}

\titlerunning{Opening Prompt Diversity for Zero- and Few-shot Keypoint Detection}

\author{Changsheng Lu$^1$\orcidlink{0000-0002-1894-286X} \and
Zheyuan Liu$^1$\orcidlink{0009-0002-3666-1778} \and
Piotr Koniusz$^{,2,1}$\orcidlink{0000-0002-6340-5289}}

\authorrunning{C.~Lu et al.}

%
\institute{$^1$The Australian National University \quad $^2$Data61/CSIRO \\
\email{changshengluu@gmail.com}, \email{zheyuan.david.liu@outlook.com}, \email{piotr.koniusz@data61.csiro.au}}

\maketitle

\begin{abstract}
  Exploiting foundation models (\eg, CLIP) to build a versatile keypoint detector has gained increasing attention. Most existing models accept either the \emph{text prompt} (\eg, ``the nose of a cat''), or the \emph{visual prompt} (\eg, support image with keypoint annotations), to detect the corresponding keypoints in query image, thereby, exhibiting either \emph{zero-shot} or \emph{few-shot} detection ability. However, the research on multimodal prompting is still underexplored, and the prompt diversity in semantics and language is far from opened. For example, how to handle unseen text prompts for novel keypoint detection and the diverse text prompts like ``Can you detect the nose and ears of a cat?'' In this work, we open the prompt diversity in three aspects: \emph{modality}, \emph{semantics} (seen \vs unseen), and \emph{language}, to enable a more general zero- and few-shot keypoint detection (Z-FSKD). We propose a novel OpenKD model which leverages a multimodal prototype set to support both visual and textual prompting. Further, to infer the keypoint location of unseen texts, we add the auxiliary keypoints and texts interpolated in visual and textual domains into training, which improves the spatial reasoning of our model and significantly enhances zero-shot novel keypoint detection. We also find large language model (LLM) is a good parser, which achieves over 96\% accuracy when parsing keypoints from texts. With LLM, OpenKD can handle diverse text prompts. Experimental results show that our method achieves state-of-the-art performance on Z-FSKD and initiates new ways of dealing with unseen text and diverse texts. The source code and data are available at \url{https://github.com/AlanLuSun/OpenKD}.

\end{abstract}  

\section{Introduction}\label{sec:intro}
%
%
Keypoint detection is a fundamental research problem in computer vision 
and has numerous applications such as pose estimation for humans~\cite{cheng2020higherhrnet,cao2019openpose,sun2019deep,yang2021transpose,fang2017rmpe,newell2016stacked} and animals~\cite{pereira2019fast,banik2021novel,cao2019cross,li2020deformation}, action recognition~\cite{wang2019comparative,mathis2018deeplabcut}, and fine-grained image classification~\cite{zhang2014part,tang2020revisiting}. 
Over the past decade, significant advancements have been made in deep keypoint detection. However, the fully-supervised methods can only predict specific body parts and species from a fixed set, which limits the model generality to novel keypoints on unseen species. The semi-supervised methods suffer from a similar problem and still require several hundreds or thousands of labels for training on new class. The unsupervised methods struggle to detect user-desired keypoints. Moreover, due to the diversity of creatures, it is difficult to collect datasets containing all species. These shortcomings necessitate such a zero- or few-shot keypoint detection: after being trained on a diverse dataset, the model can quickly recognize novel or base keypoints in unseen species given only \emph{zero, one or a few labelled samples}. Such a quick adaptation to new task mimics the transfer ability of humans, which is crucial in versatile keypoint detection.

On the one hand, researchers have recently explored few-shot keypoint detection (FSKD)~\cite{lu2022few,lu2024detect,lu2023saliency,yao2021one}. The FSKD inspired by few-shot learning~\cite{snell2017prototypical,sung2018learning,finn2017model,vinyals2016matching,koch2015siamese} is general and offers higher flexibility of detecting a varying number of keypoints in a query image, given the prompts formed by support image with keypoint annotations. In this regard, the support set is namely visual prompt, which is required when evaluating new classes. On the other hand, in the era of foundation model, the vision-language model (\eg, CLIP~\cite{radford2021learning}) injects new life to detecting keypoints, enabling zero-shot keypoint detection (ZSKD)~\cite{zhang2023clamp,zhang2023language}. Compared to FSKD, ZSKD does not need support image with annotations, instead, using the text prompt to instruct the model to detect the specified keypoints in query image. In this case, the support set is text prompt. 
Though impressive for the pioneering FSKD and ZSKD models, we identify an important issue limiting the progress: \emph{the prompt diversity for current keypoint detection methods is far from opened, especially in the aspects of modality, semantics, and language.}


\begin{figure}[!tb]
  \centering
  \includegraphics[width=\linewidth]{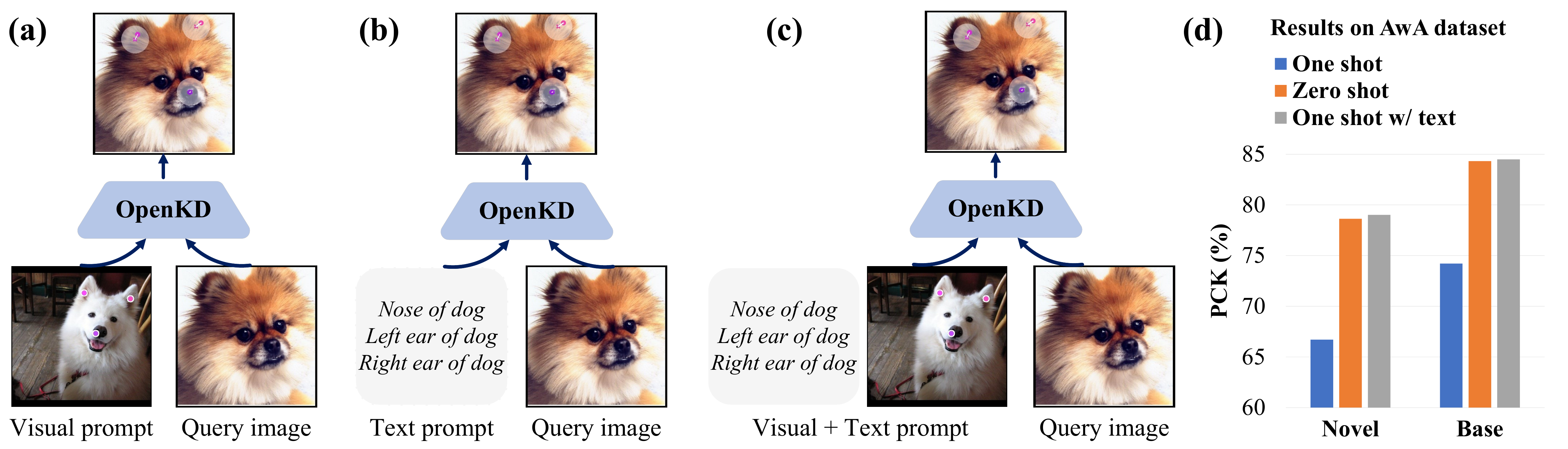}
  \caption{Illustration of multimodal prompting for keypoint detection. Our model can successfully detect keypoints given visual prompts formed by support images and keypoints (a), text prompts (b), or both (c). Graph (d) shows our model well combines the advantages of different modalities, mitigating the weakness induced by either modality.
  }
  \label{fig:intro-multi-modal}
\end{figure}

\textbf{Modality diversity.} Most existing keypoint detection models cannot support multimodal prompts, \eg, image, text, or both (see Fig.~\ref{fig:intro-multi-modal}). Multimodal prompting is more friendly in real-world interaction, and coherent with the human concept recognition. We not only see the objects, but also describe objects with language, which ultimately renders a deeper understanding of object concepts. Our work extends existing FSKD and ZSKD, building a more general zero- and few-shot keypoint detection by leveraging a multimodal prototype set and aligning the visual keypoint features towards the textual features. While straightforward, our method enables one to study the advantages of respective modality data and exploit them for better model training and testing.




\textbf{Semantic diversity.} Considering keypoints between seen and unseen species, there exist large similarity yet difference in semantics. A strong advantage of text prompt is that the keypoints with the same semantics share high similarity in language across species, which enables excellent ZSKD on base keypoints. However, if the text has different semantics, significant domain gap arises. For instance, ``the eye of a cat'' \vs ``the eye of a dog'' has cosine similarity of 0.93 using CLIP text embeddings, whereas for ``the eye of a cat'' \vs ``the knee of a dog'' it drops to 0.77. We observe that the keypoint detectors cannot perform well if text is unseen during training. To bridge the gap, we propose to open the semantic diversity by adding auxiliary keypoints and texts into training, where the auxiliary texts are reasoned by LLM given the base keypoints. To ease the reasoning error, we ask LLM a good question using chain of thought~\cite{wei2022chain}. We also propose a novel sampling strategy with false text control to improve the matching between auxiliary keypoints and texts. With them together, our model significantly improves ZSKD on novel keypoints. 
%
%

\begin{figure}[!tb]
  \centering
  \includegraphics[width=\linewidth]{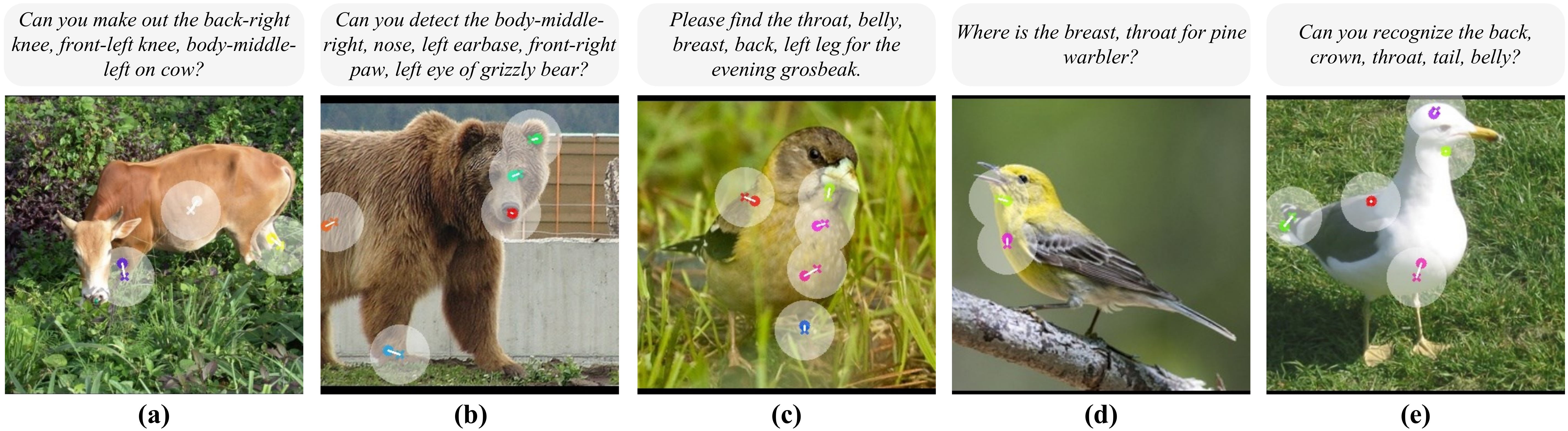}
  \caption{Examples of keypoint detection under diverse text prompting. With LLM, our method can deal with diverse texts, showing potential for real-world applications. The circles and crosses refer to predictions and GT, respectively. A keypoint is regarded as a correct detection if falling in the white shadow area that signifies PCK@0.1.
  }
  \label{fig:intro-diverse-text}
\end{figure}

\textbf{Language diversity.} Following zero-shot image classification by CLIP~\cite{radford2021learning}, existing ZSKD also constructs \emph{simple text prompts} based on templates, \eg, ``$\langle$keypoint$\rangle$'' or ``the $\langle$keypoint$\rangle$ of a $\langle$category$\rangle$ in the photo'', where $\langle\cdot\rangle$ is replaced by class names. In real-world interaction, humans tend to question in diverse styles, which results in \emph{diverse text prompts}. For example, ``Please locate the right-back leg on cat'', ``Can you find the left eye, right ear and nose in image?'', \etc. A natural question is how to handle the diverse text prompting? To address this, we propose a simple yet effective approach, whose key idea is to borrow the large language model (LLM) such as GPT3.5~\cite{brown2020language} or Vicuna~\cite{chiang2023vicuna,zheng2023judging} to parse the diverse texts via prompt engineering. After extracting the keypoint type and object category, they will be synthesized into simple text prompts to instruct the detection model. As such, the diverse text prompting can be transformed into simple text prompting, which sheds light on opening language diversity (Fig.~\ref{fig:intro-diverse-text}). 

In summary, in this paper, we propose an OpenKD model with several intriguing features: 1) supporting both visual and textual prompts, 2) having the potential to handle unseen texts and diverse texts, and 3) maintaining strong generality and performance on ZSKD and FSKD. We report that LLM is capable of being a reasoner for text interpolation, and a good language parser for parsing diverse texts. We also contribute four diverse text prompt sets for the popular Animal pose dataset~\cite{cao2019cross}, AwA~\cite{banik2021novel}, CUB~\cite{WahCUB_200_2011}, and NABird~\cite{van2015building} for fair evaluations. To our best knowledge, we are the first to open semantics and language diversity of text prompts for keypoint detection.


%

\section{Related Work}\label{sec:related_work}
\noindent\textbf{Keypoint Detection} has been widely studied ranging from the traditional interest points~\cite{lowe2004sift,derpanis2004harris} to deep corner detection~\cite{zhao2023deep}, semi-supervised~\cite{moskvyak2021semi,wang2022pseudo,honari2018improving} and fully-supervised keypoint estimation~\cite{tompson2014joint,newell2016stacked,cao2019openpose,cheng2020higherhrnet,fang2017rmpe,sun2019deep,yang2021transpose,xu2022vitpose}. 
There are two major classes of methods for deep keypoint localization: i) direct coordinates based regression~\cite{carreira2016human,toshev2014deeppose} and ii) heatmap based regression. 
In contrast to existing heatmap based models dedicated to specific body parts, \eg, top-down~\cite{sun2019deep,fang2017rmpe,he2017mask} and bottom-up pose estimators~\cite{newell2016stacked,cao2019openpose,cheng2020higherhrnet}, our OpenKD model offers more flexible keypoint detection, breaking the limitation of keypoint types to be detected.

\noindent\textbf{Few-shot Keypoint Detection} 
%
is more versatile and data-efficient compared to the supervised paradigms. Similar to standard few-shot learning (FSL) \cite{vinyals2016matching}, FSKD also takes episodes for training and evaluation. Many well-known FSL methods such as ProtoNet~\cite{snell2017prototypical}, RelationNet~\cite{sung2018learning}, LwoF~\cite{gidaris2018dynamic} and MAML~\cite{finn2017model} have been extended to the field of keypoint detection~\cite{lu2022few,bohdal2023meta,ge2021metacloth} and serve as baselines. Recently, Lu and Koniusz \cite{lu2022few} formalize the comprehensive FSKD settings and model the localization uncertainty of keypoints. Further, they propose a lightweight FSKD model~\cite{lu2024detect} and explore FSKD under transductive setting and occlusions~\cite{lu2023saliency}. 
Bohdal \etal~\cite{bohdal2023meta} propose a dataset-of-datasets, benchmarking the FSL algorithms universal to various vision tasks. 
FSKD also inspires class-agnostic animal pose estimation~\cite{lu2022few,lu2024detect,xu2022pose}. Compared to existing FSKD, our work pushes further by opening the text prompting, which enables more versatile keypoint detection. Moreover, text prompts can complement visual prompts, delivering better performance of detecting base keypoints on unseen species.



\noindent\textbf{Zero-shot Keypoint Detection} 
%
becomes feasible thanks to the great progress of vision-language pre-training (\eg, CLIP~\cite{radford2021learning} and BLIP~\cite{li2022blip}). 
Since CLIP was pre-trained on internet-scale image-text pairs that contain more semantic concepts than other datasets, CLIP can provide semantically rich features for various downstream tasks~\cite{mukhoti2023open,zhong2022regionclip,zhang2023clamp}, showing strong zero-shot transfer ability. 

Recently, CLAMP~\cite{zhang2023clamp} adapts CLIP for animal pose estimation in feature-aware and spatial-aware aspects; Approach~\cite{zhang2023language} also employs CLIP as backbone and proposes language-driven keypoint detection. Compared to FSKD which still requires one or more annotations, ZSKD merely demands cheap language descriptions, which make it intriguing for keypoint detection. Nevertheless, most existing ZSKD works cannot support multimodal prompts and their prompt diversity is limited. To address these issues, we propose an OpenKD model which further opens modality, semantics, and language. To handle unseen texts, we propose to generate auxiliary keypoint-text pairs which significantly help reason novel keypoints on unseen species. Moreover, we contribute a simple yet effective language parsing module to handle diverse text prompts.



\noindent\textbf{Foundation Models} 
%
%
%
include a broad set of models pre-trained over large-scale datasets, including LLMs (\eg,  GPT3.5~\cite{brown2020language}), VLMs (\eg, CLIP \cite{radford2021learning}), and Multimodal models (\eg,  GPT4~\cite{achiam2023gpt}). Transferring the knowledge from these foundation models to specific tasks \emph{in a cost-effective way} is popular in computer vision~\cite{jiao2024toward,liu2023grounding,zhang2023language,zhang2023clamp,lu2020deep}. Thus, we follow this trend and leverage LLM like GPT3.5 (relatively cheap) to perform reasoning and parsing for keypoints.


\section{Method}\label{sec:method}

\subsection{Model Architecture}
%
Following the general few-shot learning~\cite{vinyals2016matching,sung2018learning,snell2017prototypical}, Z-FSKD model is evaluated on episodes, each of which includes a support set and a query set. The query set is comprised of query images, while the support set gives the prompts. If the \emph{visual prompts}, \ie, $K$ support images with keypoint annotations ($K\geq 1$), are given, then the problem is defined as \emph{$K$-shot detection}. If only the \emph{text prompt} (\ie the language description) is given, then $K=0$ thus the problem becomes \emph{zero-shot detection}. The goal of Z-FSKD is to detect the corresponding keypoints in query image given the prompts, whether visual prompt, text prompt, or both. Such an approach allows the model to effectively respond to diverse inputs.

\begin{figure}[!tb]
  \centering
  \includegraphics[width=\linewidth]{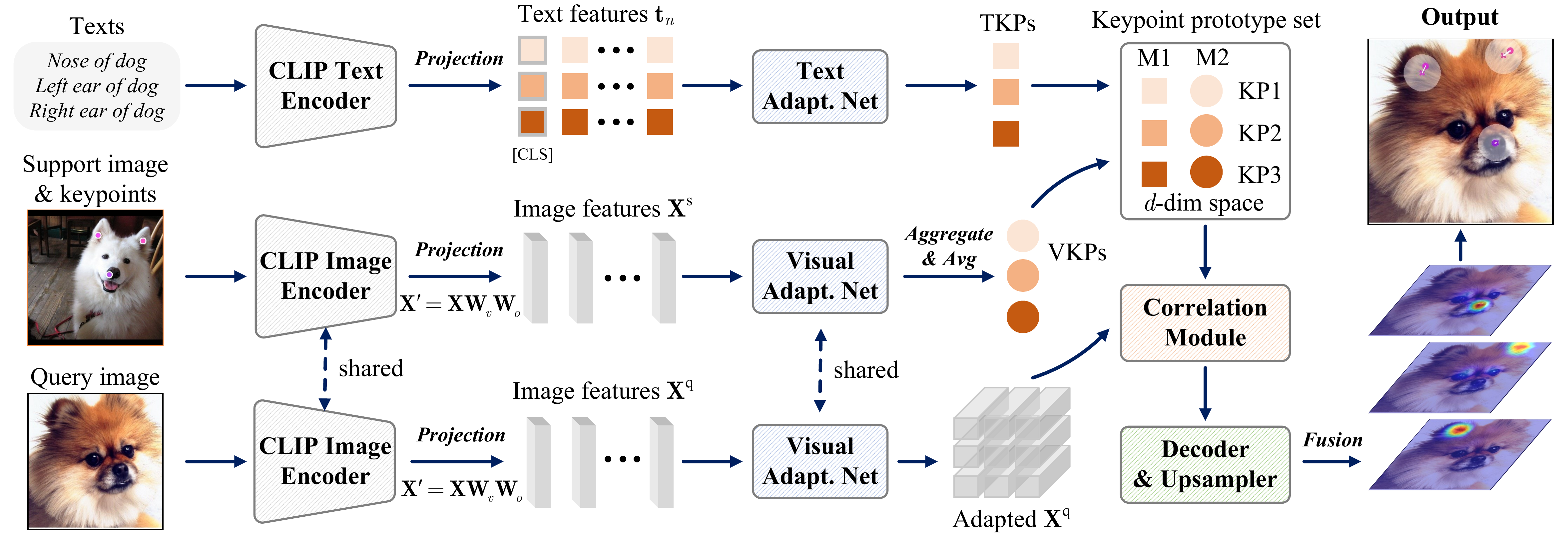}
  \caption{The sketch of model inference. Our OpenKD allows testing under visual prompt, text prompt, or both. For clarification, we show the ``both'' case (\ie, 1-shot with text testing). We firstly extract the deep features of texts, support and query images via CLIP, and then adapt both modalities of features via residual refinement. After extracting the visual keypoint prototype (VKP) and textual counterpart, we build the prototype set to perform class-agnostic correlation and heatmap decoding. Finally, we fuse the heatmaps induced by two modalities (\ie, M1 \& M2) to obtain predictions.
  }
  \label{fig:pipeline}
\end{figure}

\begin{figure}[!tb]
  \centering
  \includegraphics[width=\linewidth]{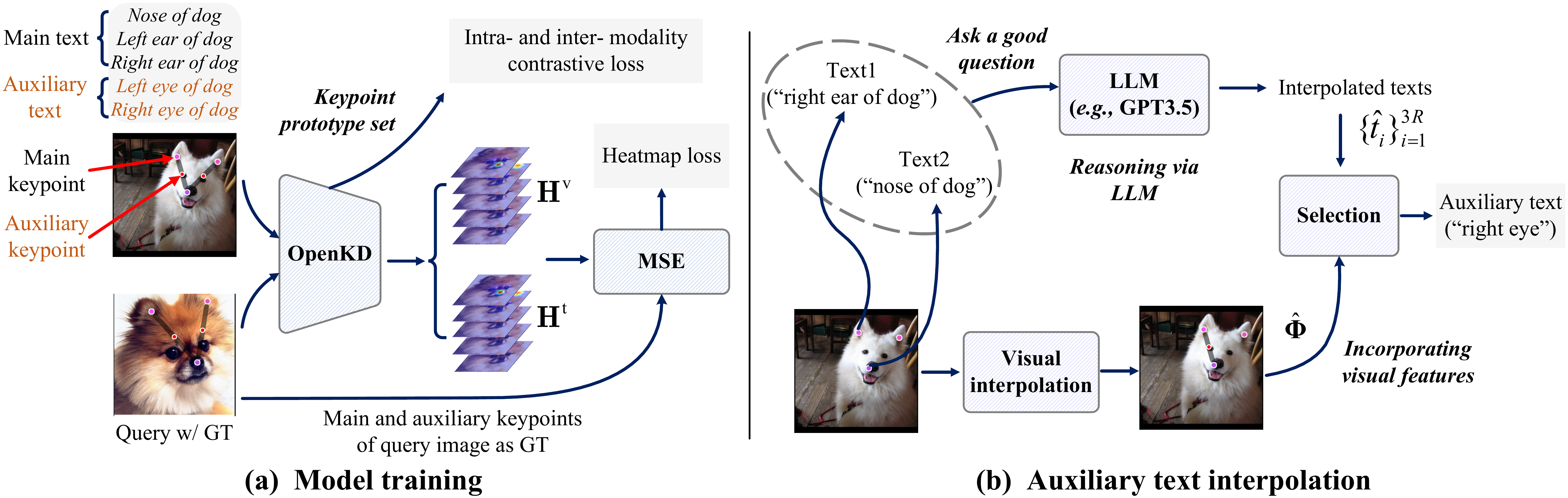}
  \caption{Model training and text interpolation. (a) In addition to multi-group heatmap regression, we improve model performance by introducing intra- and inter-modality contrastive learning and the novel auxiliary keypoint and text learning. (b) We exploit LLM for auxiliary texts interpolation and explore incorporating visual keypoint features for text selection in order to mitigate the noisy texts or false texts.
  }
  \label{fig:training-and-itpl}
\end{figure}

The overview of our model inference is shown in Fig.~\ref{fig:pipeline}, which mainly includes four stages: i) image/text feature extraction, ii) feature adaptation, iii) keypoint prototype set building, iv) correlation, decoding and heatamp fusion. 
The overview of our model training is shown in Fig.~\ref{fig:training-and-itpl}, which provides the novel approaches to improve model performance. 


\noindent\textbf{Image/Text Feature Extraction}\label{sec:arch-feature-extraction} 
Since the input of our model involves vision and language two different modalities, we wish the features extracted from image and text have smaller modality gap, so that the text can somewhat correlate with image regions of the same semantics. Such a property could help the model find keypoint locations in query image. To this end, we resort to the CLIP~\cite{radford2021learning} which is pre-trained on large-scale image-text pairs. The CLIP includes image and text encoders, where the text encoder is generally a transformer~\cite{vaswani2017attention} while the image encoder can be CNN~\cite{lecun2015deep} or ViT~\cite{dosovitskiy2020image}. 
We empirically found that the CLIP image encoder based on RN50~\cite{radford2021learning,he2016deep}, a CNN model, gives high efficiency in consideration of both performance and cost. 
Thus, we choose it as the default backbone. However, when extracting the image features, original CLIP image encoder only retains the classification token $\mathbf{x}_{\text{class}}$ and discards image tokens $\mathbf{X}$ by \emph{attention pooling} as
\begin{equation}
  \mathbf{x}_{\text{class}} = \text{AvgPool}(\mathbf{X}), \quad \mathbf{x}_{\text{class}}' = \text{Attention}(\mathbf{x}_{\text{class}}\mathbf{W}_q, \mathbf{X}\mathbf{W}_k, \mathbf{X}\mathbf{W}_v)\mathbf{W}_o.
\end{equation}
While it is natural for image-text matching, we require the image tokens to recover the spatial locations of keypoints. Thus, we propose to obtain the image tokens via a projection composed by $\mathbf{W}_v$ and $\mathbf{W}_o$ as
\begin{equation}
  \mathbf{X}' = \mathbf{X}\mathbf{W}_v\mathbf{W}_o.
  \label{eq:projection}
\end{equation}
In this way, we can not only handle the \emph{channel inconsistency} between raw image tokens $\mathbf{X}$ and text features, but also \emph{reuse} the projection layers $\mathbf{W}_v$ and $\mathbf{W}_o$. 

For brevity, we assume that per input episode contains one query image $\mathbf{I}^\text{q}$, one support image $\mathbf{I}^\text{s}$ with $N$ annotated keypoints, and $N$ keypoint texts. With Eq.~\ref{eq:projection}, both the support and query images can be encoded and projected as the support and query feature maps as $\mathbf{X}^\text{s} = \mathcal{F}_{\text{v}}(\mathbf{I}^\text{s})$ and $\mathbf{X}^\text{q} = \mathcal{F}_{\text{v}}(\mathbf{I}^\text{q})$ in deep feature space $\mathbb{R}^{l \times l \times d}$ via a shared CLIP image encoder $\mathcal{F}_{\text{v}}$. Further, with the CLIP text encoder $\mathcal{F}_{\text{t}}$, the $N$ texts are firstly tokenized, then encoded as the text features $\mathbf{t}_n \in \mathbb{R}^{m \times d}$, where $n=1,2,\cdots, N$; and $m$ is sequence length. 


\vspace{0.1cm}
\noindent\textbf{Feature Adaptation}\label{sec:arch-feature-adapt} 
Despite CLIP can provide general yet transferrable features, we need to adapt them into a new multimodal feature space that is more suitable for keypoint detection. 
Inspired by residual learning~\cite{he2016deep}, we propose two adaptation nets $\mathcal{A}_{\text{v}}$ and $\mathcal{A}_{\text{t}}$ to respectively refine the obtained image and text features in a residual way:
\begin{equation}
  \mathbf{X}^{\text{s}} := \mathbf{X}^{\text{s}} + \mathcal{A}_{\text{v}}(\mathbf{X}^{\text{s}}), \quad \mathbf{X}^{\text{q}} := \mathbf{X}^{\text{q}} + \mathcal{A}_{\text{v}}(\mathbf{X}^{\text{q}}), \quad \mathbf{t}_{n} := \mathbf{t}_{n} + \mathcal{A}_{\text{t}}(\mathbf{t}_{n}).
\end{equation}
The adaptation net $\mathcal{A}$ is highly scalable and can choose transformer~\cite{vaswani2017attention} or bottleneck~\cite{he2016deep} as refinement blocks. 
We found that $\mathcal{A}$ can well adapt features. 


\vspace{0.1cm}
\noindent\textbf{Keypoint Prototype Set}\label{sec:arch-prototype-set} 
Once we obtain the adapted features of support image and text prompts, \ie, $\mathbf{X}^{\text{s}}$ and $\mathbf{t}_{n}$, we propose to convert them into keypoint prototypes, which are unified in a keypoint prototype set. 

For visual prompt, recall that keypoint labels are provided along with support image. Thus, we can aggregate the the local features for each support keypoint $\mathbf{p}_{n}$ by using pixelwise weighted summation between feature map $\mathbf{X}^{\text{s}}$ and Gaussian heatmap $\mathbf{H}(\mathbf{p}_n;\sigma)$, yielding the visual keypoint representation (VKR) as $\mathbf{\Phi}_{n} \in \mathbb{R}^{d}$. The $\sigma$ is the standard deviation that controls Gaussian spread. If $K$ support images are given, namely in few-shot case, the VKRs of the same type of keypoints $\mathbf{\Phi}_{k,n}$, will be averaged to build the visual keypoint prototype (\textbf{VKP}) $\mathbf{\Psi}^{\text{v}}_{n} = \frac{1}{K}\sum_{k} \mathbf{\Phi}_{k,n}$. Analogously, we can develop textual keypoint prototype (\textbf{TKP}) $\mathbf{\Psi}^{\text{t}}_{n}$ by text features if provided with multiple texts per keypoint. Finally, we can build the keypoint prototype set as $\mathcal{T}=\mathcal{T}^{\text{v}} \cup \mathcal{T}^{\text{t}}$, where $\mathcal{T}^{\text{v}}=\{\mathbf{\Psi}^{\text{v}}_{n}\}$ is the VKPs and $\mathcal{T}^{\text{t}}=\{\mathbf{\Psi}^{\text{t}}_{n}\}$ is the TKPs. 
In this way, different-modal prompts can be summarized as keypoint prototypes in a shared $d$ dimensional feature space, which enables our model to flexibly handle various modalities. 
Moreover, each prototype in set $\mathcal{T}$ can guide the model to induce a keypoint heatmap, establishing the zero- and few-shot keypoint detection.


\vspace{0.1cm}
\noindent\textbf{Correlation, Decoding and Fusion}\label{sec:arch-correlation-decoder} 
To discover the corresponding keypoints in query image $\mathbf{I}^\text{q}$, each keypoint prototype $\mathbf{\Psi}_{n}$, either VKP $\mathbf{\Psi}^{\text{v}}_{n}$ or TKP $\mathbf{\Psi}^{\text{t}}_{n}$, is required to be correlated with the query feature map $\mathbf{X}^{\text{q}}$. To this end, we leverage a correlation module $\mathcal{C}$ which takes both $\mathbf{\Psi}_{n}$ and $\mathbf{X}^{\text{q}}$ as input to produce the attentive feature maps $\mathbf{A}_{n} = \mathcal{C}(\mathbf{\Psi}_n, \mathbf{X}^{\text{q}})$, where $\mathbf{A}_{n} \in \mathbb{R}^{l \times l \times d}$ and $n=1,2,\cdots,N$. 
For the correlation module $\mathcal{C}$, we explored multiple variants and found that the simple \emph{cross correlation} already yields high performance, \ie, $\mathbf{A}_{n} = \mathbf{X}^{\text{q}} \odot \mathbf{\Psi}_n$, where $\odot$ denotes the channel-wise multiplication.

Subsequently, a \emph{class-agnostic} decoder $\mathcal{D}$ is devised to convert each attentive feature map into a keypoint heatmap, \ie, $\mathbf{H}_{n} = \mathcal{D}(\mathbf{A}_{n})$, where $\mathbf{H}_{n} \in \mathbb{R}^{l \times l}$. For heatmap regression based keypoint localization, the higher resolution of heatmap can greatly reduce the coordinate decoding error. Thus, a lightweight upsampling module $\mathcal{U}$ is adopted to further refine heatmaps, \ie $\mathbf{H}_{n} := \mathcal{U}(\mathbf{H}_{n})$. 

Since each modality of prototypes can induce one group of heatmaps, considering visual and textual modalities, we have two groups of heatmaps $\mathbf{H}^{\text{v}} = \{\mathbf{H}^{\text{v}}_{n}\}_{n=1}^{N}$ and $\mathbf{H}^{\text{t}} = \{\mathbf{H}^{\text{t}}_{n}\}_{n=1}^{N}$. In testing phase, we \emph{fuse} the upsampled multi-group heatmaps as the final output $\mathbf{H} \in \mathbb{R}^{N \times ul \times ul}$ ($u$ is upsampling factor):
\begin{equation}
    \mathbf{H} = (\mathbf{H}^{\text{v}} + \mathbf{H}^{\text{t}}) / 2
\end{equation}
During training phase, inspired by HigherHRNet~\cite{cheng2020higherhrnet}, we perform \emph{group-specific supervision} for heatmaps as
\begin{equation}
    \mathcal{L}_{\text{kp}} = (\|\mathbf{H}^{\text{v}}-\mathbf{H}^{*}\|_2^2 + \|\mathbf{H}^{\text{t}}-\mathbf{H}^{*}\|_2^2) / 2
    \label{eq:kp-loss}
\end{equation}
where $\mathbf{H}^{*}$ denotes the groundtruth heatmap that encodes query keypoint.


\vspace{0.1cm}
\noindent\textbf{Improve Model by Contrastive Learning}\label{sec:arch-contrastive-learning} 
Since the keypoint prototype directly impacts performance, we aim for its representation ability to be as strong as possible. The same type of keypoints should have higher invariance across species, \eg, the ear of cat, dog, and cow; while different types are required to be sufficiently discriminative such that our model can distinguish the ambiguous texts or fine-grained texts like ``the left ear'' \vs ``the right ear''. To this end, we propose to introduce contrative loss over TKPs. If randomly sampling \emph{two} species $(s, s')$ at a time and each species pertaining to an episode, we have pairwise sets of TKPs, \ie, $\mathcal{T}^{\text{t}}_{s}\!=\!\{\mathbf{\Psi}_{n}\}$ and $\mathcal{T}^{\text{t}}_{s'}\!=\!\{\mathbf{\Psi}'_{n}\}$, 
which can form similarity matrix as 
\begin{equation}
    \mathbf{J}(\mathcal{T}^{\text{t}}_{s}, \mathcal{T}^{\text{t}}_{s'}) = \left(
        \begin{array}{ccc}
            \cos(\mathbf{\Psi}_1, \mathbf{\Psi}'_1) & \cdots & \cos(\mathbf{\Psi}_1, \mathbf{\Psi}'_N) \\
            \vdots &  & \vdots \\ 
            \cos(\mathbf{\Psi}_N, \mathbf{\Psi}'_1) & \cdots & \cos(\mathbf{\Psi}_N, \mathbf{\Psi}'_N) \\
        \end{array}
    \right) 
    \label{eq:similarity}
\end{equation}
where $\cos(\cdot, \cdot)$ denotes the cosine similarity. Then, the contrastive loss within textual modality $\mathcal{L}_{\text{tt}}$ becomes
\begin{equation}
    \mathcal{L}^{s \rightarrow s'}_{\text{tt}} = -\langle \mathbb{I},  \log(\text{softmax} (\mathbf{J}(\mathcal{T}^{\text{t}}_{s}, \mathcal{T}^{\text{t}}_{s'}) / \tau))\rangle,
    \quad
    \mathcal{L}_{\text{tt}} = \frac{1}{2} (\mathcal{L}^{s \rightarrow s'}_{\text{tt}} + \mathcal{L}^{s' \rightarrow s}_{\text{tt}})
    \label{eq:cl-tt-pairwise}
\end{equation}
where $\mathbb{I}$ is identity matrix and $\langle\cdot,\cdot\rangle$ denotes inner product. Since CLIP is trained via \emph{image-level alignment} instead of \emph{keypoint-level alignment} to text, we wish VKPs and TKPs are aligned so that we can exploit their respective advantages to help each other. This motivates us to propose the second contrastive loss between VKPs and TKPs. Similar to Eq.~\ref{eq:cl-tt-pairwise}, we build $\mathcal{L}_{\text{vt}}$. We find that visual-textual alignment improves 1-shot testing, but unintentionally drops 0-shot performance slightly. We discover the principle behind this is due to the better clustering effects of textual keypoint representations than visual ones. To address this issue, we perform \emph{stop gradient} to TKPs in $\mathcal{L}_{\text{vt}}$, thus enforcing the VKPs to align towards TKPs. We also find adding $\mathcal{L}_{\text{vv}}$ does not further advance scores. We conjecture the reason is after aligning VKPs to TKPs via $\mathcal{L}_{\text{vt}}$, the visual features will also improve discrimination along with textual features as $\mathcal{L}_{\text{tt}}$ is applied.

\vspace{0.1cm}
\noindent\textbf{Optimization}\label{sec:arch-optimization} 
By incorporating the heatmap regression loss $\mathcal{L}_{\text{kp}}$, intra- and inter-modality contrastive loss $\mathcal{L}_{\text{tt}}$ and $\mathcal{L}_{\text{vt}}$, we have the overall loss as
\begin{equation}
    \mathcal{L} = \lambda_{1}\mathcal{L}_{\text{kp}} + \lambda_{2}\mathcal{L}_{\text{tt}} + \lambda_{3}\mathcal{L}_{\text{vt}}
    \label{eq:overall-loss}
\end{equation}
where $\lambda_i$ is the loss weight. Note that contrastive losses serve as a regularization for the main task loss. By default, we set $\lambda_{1}=1$, $\lambda_{2}=\lambda_{3}=0.002$.  


\subsection{Training with Auxiliary Keypoints and Texts}
%
%
%
Localizing the seen keypoints and texts is relatively easy to handle, but it would be difficult if the unseen ones are encountered. Thus, we propose to add auxiliary keypoints and texts interpolated from visual and textual domains into training, which significantly improves the ability to detect novel keypoints.


\vspace{0.1cm}
\noindent\textbf{Visual Interpolation}\label{sec:itpl-keypoint} 
Following~\cite{lu2022few}, we generate the auxiliary keypoints in visual domain via interpolation $\mathcal{I}^{\text{v}}(z; \mathbf{p}_1, \mathbf{p}_2)$, where $z\in(0,1)$ is the interpolation node and $\mathbf{p}_i \in \mathbb{R}^2$ is the end points that form an interpolation path. An off-the-shelf saliency detector~\cite{wu2019stacked} is adopted to filter the auxiliary keypoints out of the foreground~\cite{lu2022few}. Via interpolation $\mathcal{I}^{\text{v}}$ in visual domain, the auxiliary keypoints $\hat{\mathbf{p}}$ can boost the visual diversity of training keypoints.


\vspace{0.1cm}
\noindent\textbf{Text Interpolation and Selection}\label{sec:itpl-text} 
The $\mathcal{I}^{\text{v}}$ can assist the model in handling unseen keypoints in visual prompt, but helps little if prompted unseen texts. Consequently, we propose using text interpolation $\mathcal{I}^{\text{t}}(z; t(\mathbf{p}_1), t(\mathbf{p}_2), c)$ to mitigate this issue, where $t(\mathbf{p}_i)$ are the names of end points $\mathbf{p}_i$ and $c$ is object category. Thanks to great language processing ability of LLM (\eg, GPT3.5~\cite{brown2020language}), we propose to ask LLM a good question to reason the auxiliary keypoint texts given base keypoints (Fig.~\ref{fig:training-and-itpl}(b)). \Eg, if we want to infer the keypoint text between ``thigh'' and ``paw'' of a ``cat'', we can design a vanilla prompt to LLM as:

\emph{``Please give me three most common body parts/keypoints at 1/2 between thigh and paw of a cat. Please answer in concise words.''} 
%
%
%
%

\noindent Apparently, we expect that the answers returned by LLM should include ``knee'' or ``elbow''. In practice, the keypoint types are usually more complex thus the reasoning is challenging. To enhance reasoning, we improve the above vanilla prompt into a more advanced one 
using chain of thought (CoT)~\cite{wei2022chain}, whose key idea is to firstly provide an example with analysis and then ask the question. We provide the detailed prompts and examples in \textbf{\S\ref{sec:suppl-itpl-text} of Suppl.} 

Considering LLM has randomness and might produce erroneous answers, we let LLM give three most possible answers for each interpolation path and repeat $R$ times, yielding a text pool with $3R$ texts, \ie, $\{\hat{t}_i\}_{i=1}^{3R}$. As such, the GT text has higher chance to be included. Afterwards, we need to select one interpolated text from the pool to form a pair with visually interpolated keypoint. To this end, a simple yet effective approach is sampling from top-1 results of all repetitions. However, it does not use the visual features for selection. An interesting approach is to use the correlation between visual and text features for selection as follows:
\begin{equation}
  j = \arg\max\nolimits_{i} \cos(\hat{\mathbf{\Phi}}, \hat{\mathbf{t}}_{i}),
  \quad
  \hat{t}^{*} = \hat{t}_{j}
  \label{eq:itpl-text-corr}
\end{equation}
where $\hat{\mathbf{\Phi}}$ is the auxiliary keypoint feature of $\hat{\mathbf{p}}$, $\hat{\mathbf{t}}_{i}$ is text feature of $\hat{{t}_i}$, and $\hat{t}^{*}$ is the selected text. However, we found that sometimes it leads performance drop. We suspect it is as CLIP's keypoint-text features are not fine-grained aligned enough on those regions beyond annotations. Opposed to picking, we observe such a correlation could be used to reject low-quality text if keypoint-text similarity $\cos(\hat{\mathbf{\Phi}}, \hat{\mathbf{t}}_{i})$ is low. Consequently, we propose a more advanced sampling with false text control (\textbf{FTC}). Namely, we sample the text within the range top-$\eta$, but the text will be rejected if the similarity $\cos(\hat{\mathbf{\Phi}}, \hat{\mathbf{t}}_{i})$ is below the threshold $\alpha$. In this way, we can form auxiliary keypoint-text pairs with higher quality to enhance model training. We will apply $\mathcal{I}^{\text{v}}$ to both support and query images to generate support and query auxiliary keypoints $\hat{\mathbf{p}}^{\text{s}}$ and  $\hat{\mathbf{p}}^{\text{q}}$, respectively. Similar to Eq.~\ref{eq:kp-loss}, $\hat{\mathbf{p}}^{\text{q}}$ encodes the GT heatmap supervising the heatmaps induced by $\hat{\mathbf{p}}^{\text{s}}$ and $\hat{t}^{*}$.

\subsection{LLM is a Good Language Parser}
%
To handle \emph{diverse text prompts} in keypoint detection, unlike current fashion of utilizing LLM for text generation, we use LLM for text decomposition. For example, if our model receives a diverse text prompt like ``Can you localize the left eye and nose of cat?'', we firstly leverage an LLM (\eg, GPT3.5~\cite{brown2020language}) to perform text parsing by providing it with a simple yet effective prompt as follows:

\textit{Please extract the animal and keypoint keywords from the below text: ``Can you localize the left eye and nose of cat?''}

\noindent Then, LLM is able to return the parsed texts regarding $\langle$keypoint$\rangle$ and $\langle$object$\rangle$, \ie, the ``left eye'', ``nose'', and ``cat'' in this example. Subsequently, we can directly leverage the parsed results to synthesize \emph{simple text prompts} to instruct our model and predicting all the keypoints corresponding to the text. In this way, the problem of diverse text prompting can be converted to simple text prompting, thus opening the language diversity for keypoint detection.


%
%


To fairly evaluate our approach, we manually collect 100 diverse text templates which cover most scenarios for questions (\textbf{\S\ref{sec:suppl-diverse-text} of Suppl.}). Take one as an example: ``Where is the $\langle$keypoint$\rangle$ for $\langle$object$\rangle$?'' Afterwards, we develop four diverse text prompt sets for popular Animal pose dataset~\cite{cao2019cross}, AwA~\cite{banik2021novel}, CUB~\cite{WahCUB_200_2011}, and NABird~\cite{van2015building}. Specifically, we accompany each object instance with a diverse text prompt synthesized by randomly sampling one template and \emph{one to $N$ valid keypoints}. Compared to traditional text prompts which only have $N$ variants, our constructed text prompt sets are more diverse, whose space could be as large as $100\cdot(C_{N}^1 + C_{N}^2 + \cdots + C_{N}^N)=100\cdot 2^N$ variants. \eg, CUB has 11 keypoints, its space is larger than $10^5$. We randomly sample 1000 diverse texts from each dataset and evaluate LLM's parsing performance. Table~\ref{tab:parsing-results} reports that GPT3.5 has over 96\% accuracy in parsing keypoints from text, which shows that LLM is a good language parser.



\section{Experiments}\label{sec:experiments}
\subsection{Experimental Settings}
\noindent\textbf{Datasets and Splits.} \textbf{1)} Animal pose dataset~\cite{cao2019cross} has five mammal species \emph{cat}, \emph{dog}, \emph{cow}, \emph{horse}, and \emph{sheep}, with over 6000 instances with keypoint annotations. Each animal species is alternately chosen as unseen species for testing while the remaining four as seen species for training, which yields \emph{five} subproblems; \textbf{2)} AwA~\cite{banik2021novel} has 35 diverse animal species with 10064 images. For AwA, 25 species are for training and 10 for testing; \textbf{3)} CUB~\cite{WahCUB_200_2011} consists of 200 species with 15 keypoint annotations. We use 100 species for training, 50 for validation, and 50 for testing; \textbf{4)} NABird~\cite{van2015building} is a larger dataset than CUB with 555 categories, 11 types of annotated body parts, and 48,562 images. The species split is 333, 111, and 111 for training, validation, and testing, respectively. To test model generalization, we follow \cite{lu2022few} to split keypoints into base and novel sets, reporting the performance of both base and novel keypoint detection on unseen species.

\noindent\textbf{Metric.} The percentage of correct keypoints (PCK) is used. A predicted keypoint is correct if its distance to GT $d\!\leq\!\rho\!\cdot\! \max ( w_{\text{bbx}}, h_{\text{bbx}} )$, where $w_{\text{bbx}}$ and $h_{\text{bbx}}$ are the edges of object bounding box. Following~\cite{lu2022few}, we set $\rho$ to $0.1$.

\noindent\textbf{Implementation Details.} The input image size for all models is $384 \times 384$. 
We freeze the CLIP text encoder while finetune last two layers of image encoder. The temperature $\tau$ used in contrastive loss is 0.05. Same to \cite{lu2022few}, the visual auxiliary keypoints are generated in pre-defined interpolation path and we set $z\!=\!0.5$. For text interpolation, to reduce randomness, we use GPT3.5 to reason $R=3$ times for all datasets except CUB with $R=10$. 
Our model is trained with 40k episodes and reports results using 1000 test episodes.

\noindent\textbf{Compared Methods.} 
For few-shot keypoint detection (FSKD), as previous works, we compare the few-shot learning models such as \emph{ProtoNet}~\cite{snell2017prototypical}, \emph{RelationNet}~\cite{sung2018learning}, and \emph{LwoF}~\cite{gidaris2018dynamic}, 
and FSKD-dedicated works \emph{FSKD-R/-D}~\cite{lu2022few}. For zero-shot keypoint detection (ZSKD), we adopt the source code of CLAMP~\cite{zhang2023clamp} and compare it in our experiments for fairness. Moreover, we build a \emph{Baseline} shared same backbone with our model but no using the specifically proposed auxiliary texts, contrastive loss, \etc. We denote our method as \emph{OpenKD}.

\begin{table*}[!tb]
  \centering
  \caption{Main results on Animal pose dataset. Each row shows the results tested on the model trained in one setting. Each PCK score is the average of five subproblems. 
  }
  \label{tab:main-results-animal}
  \resizebox{\textwidth}{!}{  
  \small
    \begin{tabular}{lcccccccccc}
    \toprule[1pt]
    \multirow{2}*{\#}&\multicolumn{4}{c}{\textbf{Training settings}} & \multicolumn{2}{c}{\textbf{1-shot testing$\;$}} & \multicolumn{2}{c}{\textbf{0-shot testing$\;$}} & \multicolumn{2}{c}{\textbf{1-shot w/ text}} \\ \cmidrule(lr){2-5}\cmidrule(r){6-7}\cmidrule(r){8-9}\cmidrule(){10-11}
     &\emph{Kp} &$\;\;$\emph{Aux. kp}$\;\;$&\emph{Text}&$\;\;$\emph{Aux. text}$\;\;$&Novel& Base &Novel & Base &Novel & Base \\\midrule[1pt]
    1&\Checkmark&               &             &                 &21.36& 50.84&  1.26&  1.93& 16.16& 32.81 \\
    2&\Checkmark&\Checkmark     &             &                 &47.54& 49.45&  2.00&  2.10& 39.24& 34.11 \\ 
    3&          &               &\Checkmark   &                 &16.18& 35.46& 25.60& 61.64& 27.02& 61.21 \\ 
    4&          &               &\Checkmark   &\Checkmark       &15.87& 31.57& 63.14& 65.31& \bfa{63.30}& \bfa{65.52} \\ 
    5&\Checkmark&               &\Checkmark   &                 &21.46& 52.15& 22.42& 61.07& 23.88& 60.49 \\ 
    \rowcolor{mylightblue}
    6&\Checkmark&\Checkmark     &\Checkmark   &\Checkmark       &\bfa{50.32}& \bfa{54.39}& \bfa{63.37}& \bfa{65.59}& 63.19& 64.93 \\ 
    \bottomrule[1pt]
    \end{tabular}
 }  
\end{table*}

\begin{table*}[!t]
  \centering
  \caption{Results on 1-shot keypoint detection for unseen species. The PCK scores on novel and base keypoint detection are reported. 
  \textbf{+T} means adding texts during testing.
  }
  \label{tab:one-shot-fskd-benchmark}
    \begin{tabular}{clccccccccc}
     \toprule[1pt]
      \multirow{2}*{\textbf{Setting}} & \multicolumn{1}{c}{\multirow{2}*{\textbf{Model}}} & \multicolumn{6}{c}{\textbf{Animal Pose Dataset}}  & \multirow{2}*{\textbf{AwA}} & \multirow{2}*{\textbf{CUB}} & \multirow{2}*{\textbf{NAB}} \\ \cmidrule(lr){3-8}
      && \textbf{Cat}  & \textbf{Dog}  & \textbf{Cow}  & \textbf{Horse} & \textbf{Sheep} & \textbf{Avg} &&& \\ \midrule[1pt]
    \multirow{7}*{Novel}  &ProtoNet &19.68 &16.18 &14.39 &12.05 &15.06 &15.47 &29.57 &51.32 &36.65 \\
                       &RelationNet &22.15 &17.19 &15.47 &13.58 &16.55 &16.99 &20.91 &56.59 &34.02 \\
                              &LwoF &22.47 &19.39 &16.82 &16.40 &16.94 &18.40 &28.54 &54.75 &34.19 \\
                            &FSKD-R &46.05 &40.66 &37.55 &38.09 &31.50 &38.77 &51.81 &77.90 &54.01 \\
                           & FSKD-D &52.36 &47.94 &44.07 &42.77 &36.60 &44.75 &64.76 &77.89 &\bfa{56.04} \\ \rowcolor{mylightblue}
                           & OpenKD &60.36 &53.58 &47.59 &49.01 &41.05 &50.32 &66.71 &\bfa{78.39} &53.35 \\ \rowcolor{mylightblue}
                & OpenKD\textbf{+T} &\bfa{69.26} &\bfa{66.81} &\bfa{62.40} &\bfa{63.21} &\bfa{54.27} &\bfa{63.19} &\bfa{79.02} &73.29 &53.40 \\
                           \midrule[0.5pt]
    \multirow{7}*{Base}   &ProtoNet &45.80 &39.83 &34.88 &35.80 &32.33 &37.73 &57.17 &80.36 &73.18 \\
                       &RelationNet &51.03 &45.85 &39.86 &41.97 &37.19 &43.18 &57.31 &79.40 &78.85 \\
                              &LwoF &50.05 &44.64 &43.47 &43.35 &37.84 &43.87 &63.87 &81.96 &81.39 \\
                            &FSKD-R &57.12 &51.12 &47.83 &49.71 &43.71 &49.90 &65.26 &87.94 &87.84 \\
                           & FSKD-D &56.38 &51.29 &48.24 &49.77 &43.95 &49.93 &66.39 &87.71 &86.99 \\ \rowcolor{mylightblue}
                           & OpenKD &63.61 &55.43 &51.18 &53.87 &47.86 &54.39 &74.22 &87.45 &85.11 \\ \rowcolor{mylightblue}
                & OpenKD\textbf{+T} &\bfa{70.47} &\bfa{65.09} &\bfa{62.84} &\bfa{66.46} &\bfa{59.81} &\bfa{64.93} &\bfa{84.50} &\bfa{91.81} &\bfa{91.23} \\
                            \bottomrule[1pt]
    \end{tabular}
\end{table*}
\begin{table*}[!t]
  \centering
  \caption{Results on 0-shot keypoint detection for unseen species. 
  CLAMP$^\dagger$ is the variant trained by adding our interpolated auxiliary texts.
  }
  \label{tab:zero-shot-fskd-benchmark}
  {
  \setlength{\tabcolsep}{1.8pt}
    \begin{tabular}{clccccccccc}
     \toprule[1pt]
      \multirow{2}*{\textbf{Setting}} & \multicolumn{1}{c}{\multirow{2}*{\textbf{Model}}} & \multicolumn{6}{c}{\textbf{Animal Pose Dataset}}  & \multirow{2}*{\textbf{AwA}} & \multirow{2}*{\textbf{CUB}} & \multirow{2}*{\textbf{NAB}} \\ \cmidrule(lr){3-8}
      && \textbf{Cat}  & \textbf{Dog}  & \textbf{Cow}  & \textbf{Horse} & \textbf{Sheep} & \textbf{Avg} &&& \\ \midrule[1pt]
    \multirow{4}*{Novel}  &Baseline &25.64&24.29&25.29&18.14&25.51&23.77&28.95&34.94&30.00\\ 
                          &CLAMP    &20.90&24.06&27.07&16.86&20.73&21.92&38.96&37.09&18.18\\
                  &CLAMP$^\dagger$  &61.70&58.09&\bfa{61.42}&64.86&53.13&59.84&77.66&69.30&50.81\\ \rowcolor{mylightblue}
                           & OpenKD &\bfa{71.07}&\bfa{66.49}&60.30&\bfa{65.53}&\bfa{53.45}&\bfa{63.37}&\bfa{78.64}&\bfa{70.16}&\bfa{52.29}\\\midrule[0.5pt]
    \multirow{4}*{Base}   &Baseline &58.69&56.62&56.58&60.00&51.35&56.65&73.45&87.12&71.25\\
                          &CLAMP    &60.71&54.04&62.58&62.73&57.30&59.47&84.16&90.97&79.10\\
                  &CLAMP$^\dagger$  &60.30&56.49&62.69&60.90&57.16&59.51&83.76&90.65&83.65\\ \rowcolor{mylightblue}
                           & OpenKD &\bfa{71.34}&\bfa{66.60}&\bfa{64.53}&\bfa{67.26}&\bfa{58.23}&\bfa{65.59}&\bfa{84.32}&\bfa{91.72}&\bfa{90.63}\\
                            \bottomrule[1pt]
    \end{tabular}
 }  
\end{table*}

\subsection{Main Results}  
\textbf{Firstly}, we explore \emph{single-modal training} and the impacts of whether or not to add the auxiliary keypoints/texts (1st-4th row, Table~\ref{tab:main-results-animal}). If only using main keypoints (1st row) or main texts (3rd row), we can observe that the models perform well on base keypoints in 1-shot (50.84\%) or 0-shot (61.64\%), but fail to detect novel keypoints. However, if the model is trained by adding the auxiliary keypoints (2nd row) or auxiliary texts (4th row), we can observe the remarkable improvements on novel keypoint detection, \eg, 47.54\% \vs 21.36\% in 1-shot testing for model trained by visual prompts (2nd row \vs 1st row); and also 63.14\% \vs 25.60\% in 0-shot testing for model trained by text prompts (4th row \vs 3rd row). The significant gains confirm the benefits of visual interpolation~\cite{lu2022few} and also highlight the effectiveness of our proposed textual interpolation. \textbf{Secondly}, we investigate \emph{multimodal training} (5th-6th row, Table~\ref{tab:main-results-animal}). As one can see, our model performs well under multimodal training (6th row), and greatly outperforms the model trained by multimodal prompts without auxiliary keypoints and texts (5th row). The higher performance is due to the fact that auxiliary keypoints and texts could boost visual and textual semantic diversity, thus enabling novel keypoint detection under 1-shot and 0-shot. \textbf{Thirdly}, we observe that 0-shot testing on base keypoints is greatly higher than 1-shot, which shows that the texts can be an excellent representation as guidance to detect keypoints. \textbf{Lastly}, we examine \emph{multimodal testing}, \ie, performing \emph{1-shot with text} testing, our model obtains 63.19\% and 64.93\% on novel and base keypoints, which highlights that our model strongly combines the advantages of both modalities, mitigating the weakness induced by either modality.

\vspace{0.1cm}
\noindent\textbf{Comparisons on FSKD and ZSKD:} We comprehensively conduct the few-shot keypoint detection across four datasets. Table~\ref{tab:one-shot-fskd-benchmark} shows our OpenKD model greatly outperforms compared methods in 15 out of 18 tasks in detecting novel or base keypoints. Moreover, if adding the texts during testing, the performance of our model can be further improved, which yields 63.19\% in Animal pose dataset (63.19\% \vs 44.75\% of prior-art FSKD-D) and 79.02\% in AwA. We also notice the improvements on CUB and NAB are modest, which might be due to the auxiliary texts are relatively harder to reason, thus bringing less bonus to few-shot model. For ZSKD (Table~\ref{tab:zero-shot-fskd-benchmark}), again, our model outperforms other methods (in 17 out of 18 tasks) and the improvement of CLAMP$^\dagger$ on novel keypoint detection strongly shows the benefits of adding auxiliary texts into training.

\begin{figure}[!tb]
  \centering
  \begin{subfigure}[b]{0.66\linewidth}
      \includegraphics[width=\linewidth]{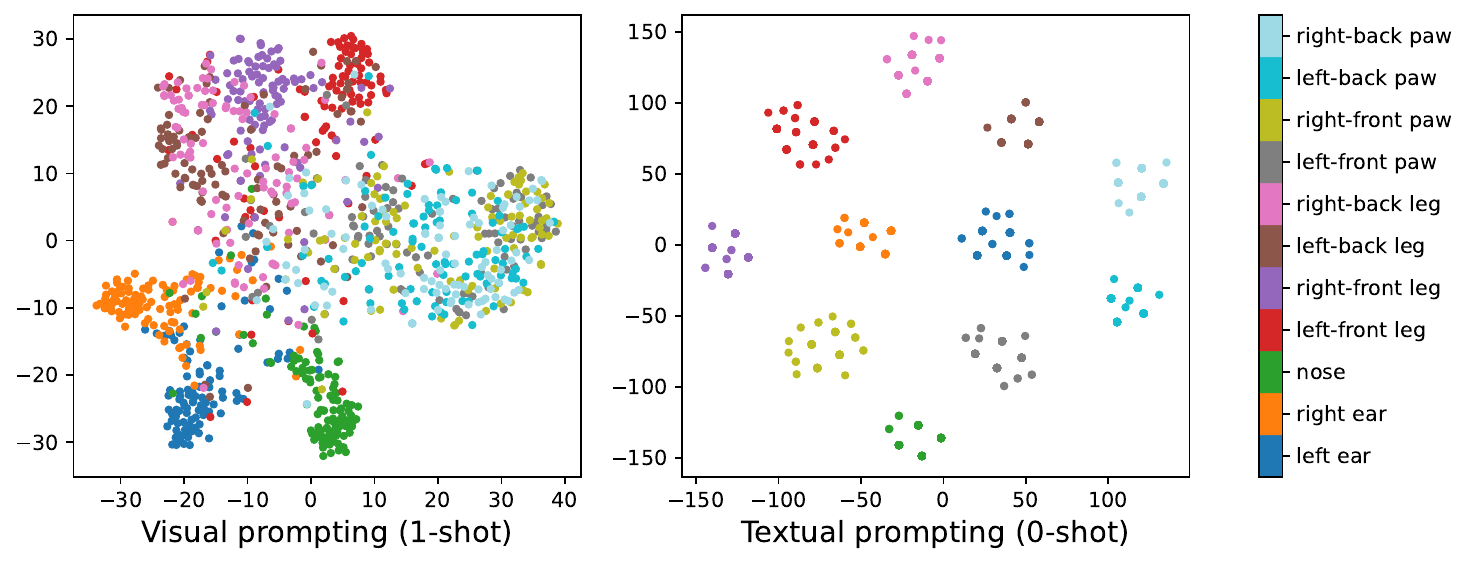}
      \subcaption{\label{fig:tsne-four2dog}}
  \end{subfigure}
  \hfill
  \begin{subfigure}[b]{0.30\linewidth}
    \includegraphics[width=.95\linewidth]{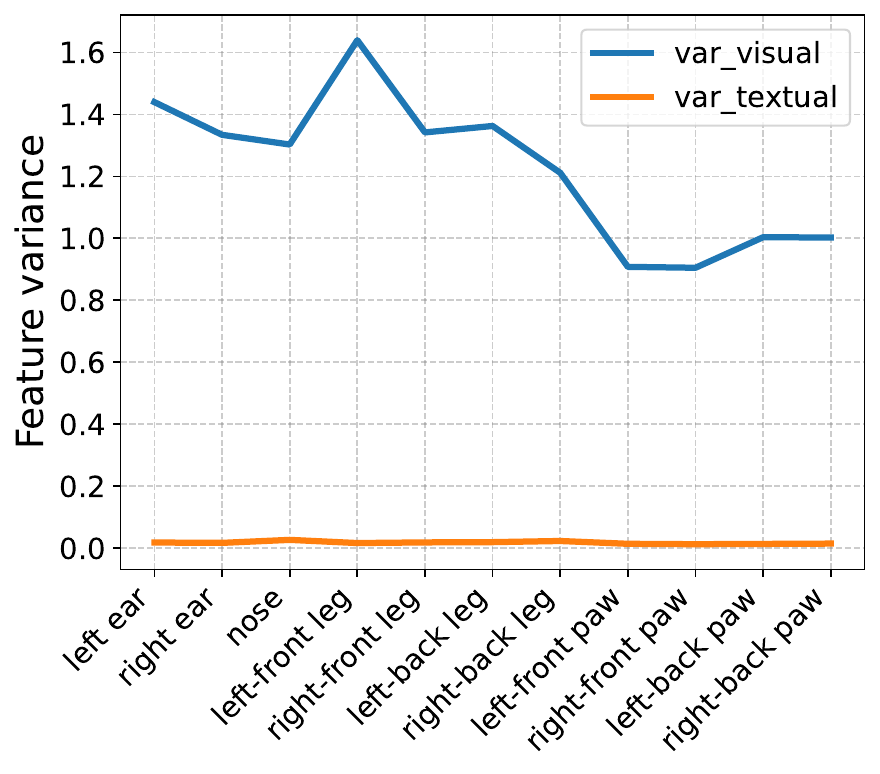}
    \subcaption{\label{fig:var-four2dog}}
  \end{subfigure}
  \caption{Different clustering effects of two modalities on base keypoints in unseen species (a) and statistical feature variance per keypoint (b).
  }
  \label{fig:tsne-var}
\end{figure}
\begin{table*}[!tb]
    \centering
    \caption{Ablation study. (a) Intra- and inter-modality contrastive loss. $\mathcal{L}_{\text{vt}}^*$ denotes no stop gradient on text features; (b) Text interpolation. Baseline$^\dagger$: w/o using auxiliary texts; CoT: chain of thought; Corr: Eq.~\ref{eq:itpl-text-corr}; FTC: sampling with false text control (ours).
    }
    \label{tab:study-cl-and-itpl-texts}
    \begin{subtable}[t]{.48\textwidth}
      \resizebox{1.0\linewidth}{!}{  
      \begin{tabular}{llcccc}
      \toprule[1pt]
      \multirow{2}*{AwA} & \multicolumn{2}{c}{1-shot} &\multicolumn{2}{c}{0-shot} \\ 
                                                                                     &Novel&Base &Novel&Base \\ \midrule[1pt]
    1: w/o CL                                                                        &65.56&72.17&76.71&81.67\\ 
    2: $\mathcal{L}_{\text{tt}}$                                                     &66.05&72.35&78.28&84.00\\ 
    3: $\mathcal{L}_{\text{tt}}$+$\mathcal{L}_{\text{vt}}^*$                         &66.70&73.91&77.10&83.85\\ \rowcolor{mylightblue}
    4: $\mathcal{L}_{\text{tt}}$+$\mathcal{L}_{\text{vt}}$                           &\bfa{66.71}&\bfa{74.22}&\bfa{78.64}&\bfa{84.32}\\ 
    5: $\mathcal{L}_{\text{tt}}$+$\mathcal{L}_{\text{vt}}$+$\mathcal{L}_{\text{vv}}$ &66.27&74.08&78.60&83.42\\ 
      \bottomrule[1pt]
      \end{tabular}
      }  
      \caption{}
      \label{tab:study-cl}
    \end{subtable}
    \begin{subtable}[t]{.49\textwidth}
      \resizebox{\linewidth}{!}{  
      \begin{tabular}{lcccc}
        \toprule[1pt]
        \multirow{2}*{0-shot testing} &\multicolumn{2}{c}{AwA} &\multicolumn{2}{c}{CUB}  \\ 
                                                 &Novel&Base &Novel&Base \\ \midrule[1pt]
        1: Baseline$^\dagger$ ($\blacktriangle $)&34.31&83.53&36.89&88.85\\ 
        2: $\blacktriangle $+w/o CoT             &73.80&82.79&68.03&88.93\\ 
        3: $\blacktriangle $+CoT                 &78.30&\bfa{84.32}&68.68&88.98\\ 
        4: $\blacktriangle $+CoT+Corr            &62.12&83.13&\bfa{71.80}&\bfa{94.27}\\ \rowcolor{mylightblue}
        5: $\blacktriangle $+CoT+FTC             &\bfa{78.64}&\bfa{84.32}&70.16&91.72\\
        \bottomrule[1pt]
      \end{tabular}
      }  
      \caption{}
      \label{tab:study-itpl-texts}
    \end{subtable}
\end{table*}

\subsection{Model and Performance Analysis} 
\noindent\textbf{Why 0-shot can outperform 1-shot?} We may observe the 0-shot testing scores on base keypoints are always higher than 1-shot testing. In this case, the training texts are accurate (\cf auxiliary text is noisy) and have high semantic similarity to testing texts. We dig more to visualize the corresponding text and keypoint features. Fig.~\ref{fig:tsne-var} shows that textual keypoint features have better \emph{clustering effects} and lower variance than visual ones, which suggests textual representations of base keypoints are better thus 0-shot yields higher scores.


\noindent\textbf{Contrastive Learning (CL).} After applying $\mathcal{L}_{\text{tt}}$, we observe the 0-shot performance significantly improves, \eg 84.00\% \vs 81.67\% (2nd row, Table~\ref{tab:study-cl}), as CL improves the discrimination between text features. However, when adding visual-textual alignment (3rd row), 1-shot testing scores boost but slightly decreasing 0-shot scores. We suspect it is due to the imbalance of representation quality between multimodal features. After adding stop gradient, it mitigates negative impacts from ``weak'' modality, thus achieving higher scores (4th row). We also observe $\mathcal{L}_{\text{vv}}$ does not further improve scores in our losses (5th row).

\begin{table*}[!tb]
    \centering
    \caption{(a) Study on repetition $R$; (b) Parsing results with GPT3.5 and Vicuna.  
    }
    \label{tab:study-repetition-and-parsing-results}
    \begin{subtable}[t]{.50\textwidth}
      \begin{tabular}{cclclclcl}
        \toprule[1pt]
        \multirow{1}*{\multirow{2}*{$R$}} & \multicolumn{4}{c}{AwA} &\multicolumn{4}{c}{CUB} \\ \cmidrule(lr){2-5}\cmidrule(lr){6-9}
                         &\multicolumn{2}{c}{Novel}        &Base & &\multicolumn{2}{c}{Novel}      &Base & \\ \midrule[1pt]
        1                &67.34 &\kern-0.3em{\tiny        }&83.98& &64.21&\kern-0.3em{\tiny       }&91.40& \\ 
        3                &78.64 &\kern-0.3em{\tiny (+11.3)}&84.32& &69.81&\kern-0.3em{\tiny(+5.60)}&92.38& \\ 
        5                &76.36 &\kern-0.3em{\tiny (+9.02)}&83.73& &69.81&\kern-0.3em{\tiny(+5.60)}&92.02& \\ 
        10               &76.61 &\kern-0.3em{\tiny (+9.27)}&83.36& &70.16&\kern-0.3em{\tiny(+5.95)}&91.72& \\ 
        \bottomrule[1pt]
      \end{tabular}
      \caption{}
      \label{tab:study-repetition}
    \end{subtable}
    \hspace{-0.3cm}
    \begin{subtable}[t]{.49\textwidth}
      \begin{tabular}{lcccc}
        \toprule[1pt]
        \multirow{2}*{Dataset}& \multicolumn{2}{c}{GPT3.5} & \multicolumn{2}{c}{Vicuna}  \\\cmidrule(lr){2-3}\cmidrule(lr){4-5}
                               &\emph{Acc. kp} &\emph{Acc. obj} &\emph{Acc. kp} &\emph{Acc. obj} \\ \midrule[1pt]
                         Animal&0.97           &0.99            &0.94           &0.98            \\ 
                            AwA&0.96           &1.00            &0.93           &0.98            \\ 
                            CUB&0.97           &0.96            &0.94           &1.00            \\ 
                         NABird&0.97           &0.99            &0.95           &1.00            \\ 
        \bottomrule[1pt]
      \end{tabular}
      \caption{}
      \label{tab:parsing-results}
    \end{subtable}
  \end{table*}
\begin{table*}[!tb]
  \centering
  \caption{Diverse text prompting evaluation. Results are the average of three runs.
  }
  \label{tab:diverse-texts-eval}
  {
    \setlength{\tabcolsep}{3pt}
  \begin{tabular}{lccccccccrr}
     \toprule[1pt]
     \multicolumn{1}{c}{\multirow{2}*{{Method}}} & \multicolumn{2}{c}{{Animal Pose}}  & \multicolumn{2}{c}{{AwA}} & \multicolumn{2}{c}{{CUB}} & \multicolumn{2}{c}{{NAB}}&\multicolumn{2}{c}{{Avg Drop}} \\ \cmidrule(lr){2-3}\cmidrule(lr){4-5}\cmidrule(lr){6-7}\cmidrule(lr){8-9}\cmidrule(lr){10-11}
&{Novel}&{Base}&{Novel}&{Base}&{Novel}&{Base} &{Novel}&{Base}&{Novel}&{Base}\\\midrule[1pt]
    OpenKD   &63.37&65.59&78.64&84.32&70.16&91.72&52.29&90.63&\multicolumn{1}{c}{-}&\multicolumn{1}{c}{-}\\
  No parsing &13.89&15.18&13.44&21.70&22.09&17.48&14.55&17.62&50.12$\downarrow$&65.07$\downarrow$\\
  Vicuna     &57.54&59.84&73.97&78.18&67.89&87.39&50.45&86.59& 3.65$\downarrow$& 5.06$\downarrow$\\ \rowcolor{mylightblue}
  GPT3.5     &61.45&62.82&76.23&80.07&69.39&90.65&52.21&89.14& \bfa{1.29}$\downarrow$& \bfa{2.39}$\downarrow$\\
    \bottomrule[1pt]
    \end{tabular}
 }  
\end{table*}

\noindent\textbf{Text Interpolation.} Table~\ref{tab:study-itpl-texts} shows that using chain of thought (CoT) significantly improves performance, especially on AwA (3rd row), as CoT improves the quality of reasoned texts. Moreover, after applying FTC, the scores can strike a high balance in all datasets (5-th row), as our FTC rejects low-quality texts and further reduces noise after including visual features into decision. We also explore the repetition $R$ for text reasoning. Table~\ref{tab:study-repetition} shows the importance of setting $R$ more than 1. 

\noindent\textbf{Diverse Text Prompting.} To evaluate our model under diverse text prompting, we randomly sample 1000 diverse texts from each dataset to conduct zero-shot testing. As shown in Table~\ref{tab:diverse-texts-eval}, after coupling with GPT3.5, our model strongly keeps the performance even under diverse texts (with less than 3\% drop). The strong results compared to no parsing shows leveraging LLM as a parser to handle language diversity is a possible solution. Moreover, we also found coupling with GPT3.5 has higher performance than Vicuna, as GPT3.5 owns stronger keypoint text parsing ability, \eg, 96\% \vs 93\% in AwA (Table~\ref{tab:parsing-results}).



\section{Conclusion}\label{sec:conclusion}
We propose to open the prompt diversity from the aspects of modality, semantics and language, to enable a more general zero- and few-shot keypoint detection. To this end, we build a versatile OpenKD model which supports both visual and textual prompting. Moreover, to bridge the semantics gap between seen and unseen texts, we propose a novel text interpolation and the selection strategy with false text control, which strongly improves zero-shot novel keypoint detection. We also discover that LLM is a good language parser. After coupling with LLM, our model can well handle diverse texts. We hope the proposed model, text interpolation and parsing approach could provide useful insights in versatile keypoint detection, thus we highly recommend it to vision community.



\section*{Acknowledgment}
Changsheng Lu is supported by Australian Government Research Training Program (AGRTP) International Scholarship. Piotr Koniusz is supported by CSIRO’s Science Digital.

%
%
\bibliographystyle{splncs04}
\bibliography{refs_camera_ready}

\clearpage

\title{OpenKD: Opening Prompt Diversity for Zero- and Few-shot Keypoint Detection (Supplementary Material)}

\titlerunning{Opening Prompt Diversity for Zero- and Few-shot Keypoint Detection}

\author{Changsheng Lu$^1$\orcidlink{0000-0002-1894-286X} \and
Zheyuan Liu$^1$\orcidlink{0009-0002-3666-1778} \and
Piotr Koniusz$^{2,1}$\orcidlink{0000-0002-6340-5289}}

\authorrunning{C.~Lu et al.}

\institute{$^1$The Australian National University \quad $^2$Data61/CSIRO \\
\email{changshengluu@gmail.com}, \email{zheyuan.david.liu@outlook.com}, \email{piotr.koniusz@data61.csiro.au}}

\maketitle

\setcounter{table}{6}
\setcounter{equation}{9}
\setcounter{figure}{5}

\appendix

Summary of the supplementary material:
\begin{itemize}
  \item \textbf{\S\ref{sec:suppl-itpl-text}}  presents more details about \textbf{how to interpolate auxiliary texts} with LLM. The chain-of-thought prompting and the examples of text interpolation are given to showcase the efficacy of the proposed method.
  \item \textbf{\S\ref{sec:suppl-diverse-text}} provides the examples of \textbf{diverse text prompts} and \textbf{text parsing}. More details about the construction of diverse text prompt sets are given.
  \item \textbf{\S\ref{sec:suppl-detail-implementation}} depicts more details about implementation.
\end{itemize}


\begin{table*}[!tb]
    \centering
    \caption{Examples of interpolating texts by prompting GPT3.5 with chain of thought on the Animal pose dataset. We use the API interface provided by OpenAI to perform reasoning. All the examples are generated automatically without human intervention.
    }
    \label{tab:supp-itpl-texts-examples}
    \begin{tabularx}{\linewidth}{lX}
      \toprule[1pt]
      \multicolumn{2}{l}{\textbf{System Instruct} You are a helpful assistant that produces keypoints of an animal.} \\\midrule
      \textbf{User}  & Q: Please give me one most common body part/keypoint at 1/2 between left-front knee and left-front paw of an animal. Please answer in concise words. Provide no excessive explanations.\newline A: The starting point is left-front knee. The end point is left-front paw. The answer should be between the starting point and end point. Left-front ankle is between the starting point and end point. The answer is left-front ankle.\newline Q: Please give me three most common body parts/keypoints at 1/2 between \textcolor{blue}{nose} and \textcolor{blue}{left ear} of \textcolor{blue}{an animal}. Pay attention to the left and right. Please answer in concise words like ``1. 2. 3.''. Please do not include \textcolor{blue}{nose} and \textcolor{blue}{left ear} in answers. Provide no excessive explanations.\\
      \textbf{GPT3.5}& 1. Left eye \textcolor{red}{$\qquad\rhd$ Correct answer, mostly desirable} \newline 2. Left cheek \textcolor{red}{$\qquad\rhd$ Correct answer, less desirable}\newline 3. Left temple \textcolor{red}{$\qquad\rhd$ Correct answer, less desirable}\\ \midrule
      \textbf{User}  & Q: Please give me one most common body part/keypoint...\newline A: The starting point is ... The end point is ... The answer should be...\newline Q: Please give me three most common body parts/keypoints at 1/2 between \textcolor{blue}{nose} and \textcolor{blue}{right ear} of \textcolor{blue}{an animal}. Pay attention to the left and right. Please answer in concise words like ``1. 2. 3.''. Please do not include \textcolor{blue}{nose} and \textcolor{blue}{right ear} in answers. Provide no excessive explanations.\\
      \textbf{GPT3.5}& 1. Right eye. \textcolor{red}{$\qquad\rhd$ Correct answer, mostly desirable}\newline 2. Temple. \textcolor{red}{$\qquad\rhd$ Correct answer, less desirable}\newline 3. Cheek. \textcolor{red}{$\qquad\rhd$ Correct answer, less desirable}\\ \midrule
      \textbf{User}  & Q: Please give me one most common body part/keypoint...\newline A: The starting point is ... The end point is ... The answer should be...\newline Q: Please give me three most common body parts/keypoints at 1/2 between \textcolor{blue}{left-front thigh} and \textcolor{blue}{left-front paw} of \textcolor{blue}{an animal}. Pay attention to the left and right. Please answer in concise words like ``1. 2. 3.''. Please do not include \textcolor{blue}{left-front thigh} and \textcolor{blue}{left-front paw} in answers. Provide no excessive explanations.\\
      \textbf{GPT3.5}& 1. Left-front knee. \textcolor{red}{$\qquad\rhd$ Correct answer, mostly desirable}\newline 2. Left-front ankle. \textcolor{red}{$\qquad\rhd$ Correct answer, less desirable}\newline 3. Left-front paw. \textcolor{red}{$\qquad\rhd$ Wrong answer, not desirable}\\ 
      \bottomrule[1pt]
    \end{tabularx}
\end{table*}

\section{Details of Text Interpolation}\label{sec:suppl-itpl-text}
%
We use the LLM to perform text interpolation $\mathcal{I}^{\text{t}}(z; t(\mathbf{p}_1), t(\mathbf{p}_2), c)$ by asking a good question, \ie, designing a good prompt to LLM. The detailed \textbf{vanilla prompt} is given as follows:

\emph{``Please give me three most common body parts/keypoints at $z$ between $t(\mathbf{p}_1)$ and $t(\mathbf{p}_2)$ of $c$. Pay attention to the left and right. Please answer in concise words like ``1. 2. 3.''. Please do not include $t(\mathbf{p}_1)$ and $t(\mathbf{p}_2)$ in answers. Provide no excessive explanations.''}

\noindent In the above formulation, $z$ is the interpolation node, $t(\mathbf{p}_i)$ is the name of starting or end point, and $c$ is the object name. In order to produce the structured output, several constraints can be appended to the prompt. For example, \emph{``Pay attention to the left and right''} could help LLM take care of the symmetry of animal anatomy. 

Although the vanilla prompt can deal with most cases, we require a more advanced prompt to improve the reasoning of LLM, as the keypoint types may be complex. The Chain of Thought (CoT)~\cite{wei2022chain} has been demonstrated to be an effective method to improve the reasoning of LLM. Inspired by CoT, we propose the \textbf{improved prompt} with CoT as follows:

\emph{
Q: Please give me one most common body part/keypoint at 1/2 between left-front knee and left-front paw of an animal. Please answer in concise words. Provide no excessive explanations.}

\emph{
A: The starting point is left-front knee. The end point is left-front paw. The answer should be between the starting point and end point. Left-front ankle is between the starting point and end point. The answer is left-front ankle.}

\emph{
Q: Please give me three most common body parts/keypoints at $z$ between $t(\mathbf{p}_1)$ and $t(\mathbf{p}_2)$ of $c$. Pay attention to the left and right. Please answer in concise words like ``1. 2. 3.''. Please do not include $t(\mathbf{p}_1)$ and $t(\mathbf{p}_2)$ in answers. Provide no excessive explanations.}

\textbf{Examples: } Table~\ref{tab:supp-itpl-texts-examples} shows three real examples of text interpolation using our improved prompt. Note that all the examples are generated at one-time run sequentially, without any human intervention, using our code. As demonstrated, our approach can successfully infer the auxiliary keypoint texts.

\begin{lstlisting}[
  % float=h,  % t, b, h
  % frame=tblr,  % tblr, none
  caption={Examples of diverse text prompts on the Animal pose dataset, AwA, CUB, and NABird. For each dataset, we show five sampled diverse texts.},
  label={lst:example-diverse-texts},
  % language=C++,  % Python
  backgroundcolor=\color{mylightgray},
  basicstyle=\ttfamily\small,  %\footnotesize
  basewidth=0.5em,
  numbers=none, %left, none
  numbersep=5pt,
  breaklines=true,
  % captionpos=b, % t or b
]
# Animal pose dataset
Please tell me the position of the left-front leg, left ear.
Can you find the left-back leg, left-front leg on cat?
Can you recognize the nose, left-front paw, left ear, left-back paw of dog?
Give me the position of the right-front knee, right-front leg.
Is there a way to pinpointing the left-back paw, right ear, right-back knee on cow?

# AwA dataset
Please pinpointing the front_right_knee, back_left_paw, left_eye, front_right_paw on ox.
Please determine the body_middle_left of otter.
The right_earbase, back_left_thai of wolf.
Give me the position of the back_left_thai, neck_base on zebra.
Please detect front_right_thai of grizzly bear.

# CUB dataset
Can you detect the crown, back, right eye, right wing of carolina wren?
Can you recognize the belly, right eye, right leg, forehead, nape, crown?
Spot the tail, right leg, back.
Where is the nape, right wing, tail, crown of a warbling vireo?
Please identify the throat, nape on brandt cormorant.

# NABird dataset
Can you locate the breast, belly, right eye, right wing, tail, nape on Bullock's Oriole?
Please detect right wing, belly, bill, right eye.
Say's Phoebe's crown, back, breast, bill, belly, left eye.
Tell me where is the back, crown, nape, right wing of California Gull?
Ovenbird's right eye, crown, belly, back.
\end{lstlisting}

\begin{table*}[!tb]
    \centering
    \caption{Examples of parsing texts via GPT3.5.
    }
    \label{tab:suppl-text-parsing-examples}
    \begin{tabularx}{\linewidth}{lX}
      \toprule[1pt]
      \textbf{User}  & Please extract the animal and keypoint keywords from the below text: ``\textcolor{blue}{Please tell me the position of the left-front leg, left ear.}''. Give the answer in simple words, like ``Animal type:, Keypoint part:''. If no animal is mentioned, set animal type to N/A.\\
      \textbf{GPT3.5}& Animal type: N/A; Keypoint part: left-front leg, left ear \textcolor{red}{$\quad\rhd$ Correct parsing}\\ \midrule
      \textbf{User}  & Please extract the animal and keypoint keywords from the below text: ``\textcolor{blue}{Can you find the left-back leg, left-front leg on cat?}''. Give the answer in simple words, like ``Animal type:, Keypoint part:''. If no animal is mentioned, set animal type to N/A.\\
      \textbf{GPT3.5}& Animal type: cat; \newline Keypoint part: left-back leg, left-front leg \textcolor{red}{$\qquad\rhd$ Correct parsing}\\ \midrule
      \textbf{User}  & Please extract the animal and keypoint keywords from the below text: ``\textcolor{blue}{Tell me where is the back, crown, nape, right wing of California Gull?}''. Give the answer in simple words, like ``Animal type:, Keypoint part:''. If no animal is mentioned, set animal type to N/A.\\
      \textbf{GPT3.5}& Animal type: California Gull; \newline Keypoint part: back, crown, nape, right wing \textcolor{red}{$\qquad\rhd$ Correct parsing}\\ 
      \bottomrule[1pt]
    \end{tabularx}
\end{table*}

\section{Details of Diverse Text Prompts}\label{sec:suppl-diverse-text}

\noindent\textbf{Examples of diverse texts:} 
A glance of diverse text prompts is shown in Listing~\ref{lst:example-diverse-texts}. For brevity, we show five randomly sampled diverse texts per dataset.

\vspace{0.1cm}
\noindent\textbf{Examples of parsing:} In our implementation, the detailed prompt to LLM for parsing is as follows:

\emph{Please extract the animal and keypoint keywords from the below text: ``$\langle\text{text}\rangle$''. Give the answer in simple words, like ``Animal type:, Keypoint part:''. If no animal is mentioned, set animal type to N/A.}

\noindent In the above example, the $\langle\text{text}\rangle$ is replaced by a specific diverse text prompt. For example, we select three diverse texts from Listing~\ref{lst:example-diverse-texts} and then use our proposed approach to parse them. Table~\ref{tab:suppl-text-parsing-examples} shows that our method can well extract the object category and keypoint texts from the diverse text. 

Moreover, in Table~\ref{tab:parsing-results} (main paper), we randomly draw 1000 diverse texts from each dataset and then perform parsing. The parsed results will be compared to groundtruth (GT). A parsed text is correct if its IoU to GT $\!\geq\!0.9$. The \emph{Acc. kp} and \emph{Acc. obj} in Table~\ref{tab:parsing-results} (main paper) refer to the parsing accuracy of keypoint/object texts. As one can see, our parsing method with GPT3.5 can achieve over 96\% accuracy in parsing keypoints from text, which shows that LLM is a good language parser.

\vspace{0.1cm}
\noindent\textbf{Diverse texts construction:} To convincingly evaluate the effectiveness of our approach, we construct \emph{four} diverse text prompt sets for the popular Animal pose dataset, AwA, CUB, and NABird. Firstly, we manually and meticulously collect 100 \textbf{diverse text templates} which cover most real-world scenarios, as shown in Listing~\ref{lst:diverse-text-templates}. Secondly, for each annotated instance in each dataset, we randomly sample one template and \emph{one to $N$ valid keypoints}. Lastly, we replace the $\langle$keypoint$\rangle$ and $\langle$object$\rangle$ (if existing) with the sampled keypoints and object category. For example, if the sampled template is ``where is the $\langle$keypoint$\rangle$ for $\langle$object$\rangle$?'', the sampled keypoints are ``left eye and right eye'', and the object category of annotated instance is ``cat'', then we can synthesize a diverse text prompt as ``where is the left eye and right eye for cat?''. We note that each annotated instance may have a varying number of valid keypoints due to self-occlusion, pose, and appearance. Thus, the synthesized text is quite diverse.

\begin{lstlisting}[
  % float=h,  % t, b, h
  % frame=tblr,  % tblr, none
  caption={Partial diverse text templates. For brevity, we show 20 templates out of 100.},
  label={lst:diverse-text-templates},
  % language=Python,
  backgroundcolor=\color{mylightgray},
  basicstyle=\ttfamily\small,  %\footnotesize
  basewidth=0.5em,
  numbers=none, %left, none
  numbersep=5pt,
  breaklines=true,
  % captionpos=b, % t or b
]
<obj>'s <keypoint>.
the <keypoint> of <obj>.
detect the <keypoint> of <obj>.
can you detect the <keypoint> of <obj>?
where is the <keypoint> for <obj>?
please detect <keypoint> of <obj>
please identify the <keypoint> on <obj>.
recognize the <keypoint> of <obj>.
please recognize the <keypoint> of <obj>.
Spot the <keypoint> of <obj>.
Locate the <keypoint> on <obj>.
please find the <keypoint> of <obj>.
distinguish the <keypoint> on <obj>.
please determine the <keypoint> of <obj>.
pinpointing the <keypoint> on <obj>.
pick out the <keypoint> on <obj>.
Could you find the <keypoint> on <obj>?
Please make out the <keypoint> on <obj>.
Give me the position of the <keypoint> on <obj>.
Please tell me the position of the <keypoint> on <obj>
\end{lstlisting}

\section{Additional Implementation Details}\label{sec:suppl-detail-implementation}
By default, the adaptation nets $\mathcal{A}_\text{v}$ and $\mathcal{A}_\text{t}$ of our OpenKD model use one bottleneck~\cite{he2016deep} and one transformer block~\cite{vaswani2017attention}, respectively. The decoder $\mathcal{D}$ uses two convolutional blocks. The upsampler $\mathcal{U}$ upscales the heatmap at a ratio of 2. 
During text sampling with false text control (FTC), we set $R=3$ and $\eta=1$ for Animal pose and AwA datasets, $R=\eta=3$ for NABird, and $R=\eta=10$ for CUB. For all datasets, we set the threshold $\alpha$ to 0.01. 
Considering the keypoint-text alignment for adapted features is weak in early training, we use 10k episodes for bootstrapping. Specifically, before 10k episodes, we adopt the original CLIP image/text features for FTC. After 10k episodes, we switch to the adapted CLIP image/text features. We note that these hyper-parameters are easy to tune and perform well in experiments.

\end{document}